\documentclass{article}

    \PassOptionsToPackage{numbers,sort&compress}{natbib}

     \usepackage[preprint]{neurips_2020}

\usepackage[utf8]{inputenc} 
\usepackage[T1]{fontenc}    
\usepackage{amsmath}
\usepackage{hyperref}       
\usepackage{url}            
\usepackage{booktabs}       
\usepackage{amsfonts}       
\usepackage{nicefrac}       
\usepackage{microtype}      
\usepackage{graphicx}
\usepackage{xcolor}
\usepackage{wrapfig}
\usepackage{subcaption}
\usepackage{tabularx}
\usepackage{appendix}
\long\def\comment#1{}

\DeclareMathOperator*{\argmin}{arg\,min}
\title{What-If Motion Prediction for Autonomous Driving}

\makeatletter
\newcommand{\printfnsymbol}[1]{%
  \textsuperscript{\@fnsymbol{#1}}%
}
\newcommand\blfootnote[1]{%
  \begingroup
  \renewcommand\thefootnote{}\footnote{#1}%
  \addtocounter{footnote}{-1}%
  \endgroup
}
\makeatother

\author{
Siddhesh Khandelwal$^*$\\
  University of British Columbia\\
  \texttt{skhandel@cs.ubc.ca} \\
   \And
   William Qi$^*$\\
   Carnegie Mellon University \\
   \texttt{wq@cs.cmu.edu} \\
   \And
   Jagjeet Singh \\
   Argo AI \\
   \texttt{jsingh@argo.ai} \\
   \And
   Andrew Hartnett \\
   Argo AI \\
   \texttt{ahartnett@argo.ai} \\
   \And
   Deva Ramanan \\
   Carnegie Mellon University and Argo AI \\
   \texttt{deva@cs.cmu.edu} \\
}

\begin{document}

\maketitle

\begin{abstract}

 Forecasting the long-term future motion of road actors is a core challenge to the deployment of safe autonomous vehicles (AVs). Viable solutions must account for both the static geometric context, such as road lanes, and dynamic social interactions arising from multiple actors. While recent deep architectures have achieved state-of-the-art performance on distance-based forecasting metrics, these approaches produce forecasts that are predicted without regard to the AV's intended motion plan.  In contrast, we propose a recurrent graph-based attentional approach with interpretable geometric (actor-lane) and social (actor-actor) relationships that supports the injection of counterfactual geometric goals and social contexts. Our model can produce diverse predictions conditioned on hypothetical or ``what-if" road lanes and multi-actor interactions. We show that such an approach could be used in the planning loop to reason about unobserved causes or unlikely futures that are directly relevant to the AV’s intended route.

\end{abstract}

\section{Introduction}
\label{introduction}

Forecasting or predicting the future states of other actors in complex social scenes is a central challenge in the development of autonomous vehicles (AVs). This is a particularly difficult task because actor futures are multi-modal and depend on other actors, road structures, and even the AV's intended motion plan. 
The emergence of large-scale AV testing, together with the public release of driving datasets and maps \cite{Argoverse, caesar2019nuscenes, lyft2019, sun2019scalability}, has stimulated promising recent work on data-driven feedforward approaches \cite{djuric2018motion, cui2019deep, bansal2018chauffeurnet, chai2019multipath, tang2019multiple, rhinehart2018r2p2} designed to address these challenges. 
\blfootnote{$^*$Denotes equal contribution}

{\bf Representations:} 
Most approaches embed both social and map information within a birds-eye-view (BEV) {\em rasterized image}, allowing learned models (typically a combination of CNNs and RNNs) to predict trajectories from extracted features. 
Although convenient, there are some drawbacks to rasterization: 1) the resulting models tend to require a large number of parameters~\cite{gao2020vectornet} and 2) some facets of the problem are best represented in coordinate spaces that are not conducive to rasterization. For example, while the physics of vehicle motion are generally modeled in Euclidean space, lane-following behaviors and map-based interactions are easier to represent in curvilinear coordinates of the road network \cite{Argoverse}. Similarly, social interactions between $N$ actors can be captured naturally in a topological graph representation with $N$ nodes; notable recent methods VectorNet \cite{gao2020vectornet} and SAMMP \cite{mercat2019multihead} take such an approach, representing individual objects as nodes that may attend to one another.

\begin{figure}[t!]
\centering
\includegraphics[width=\linewidth]{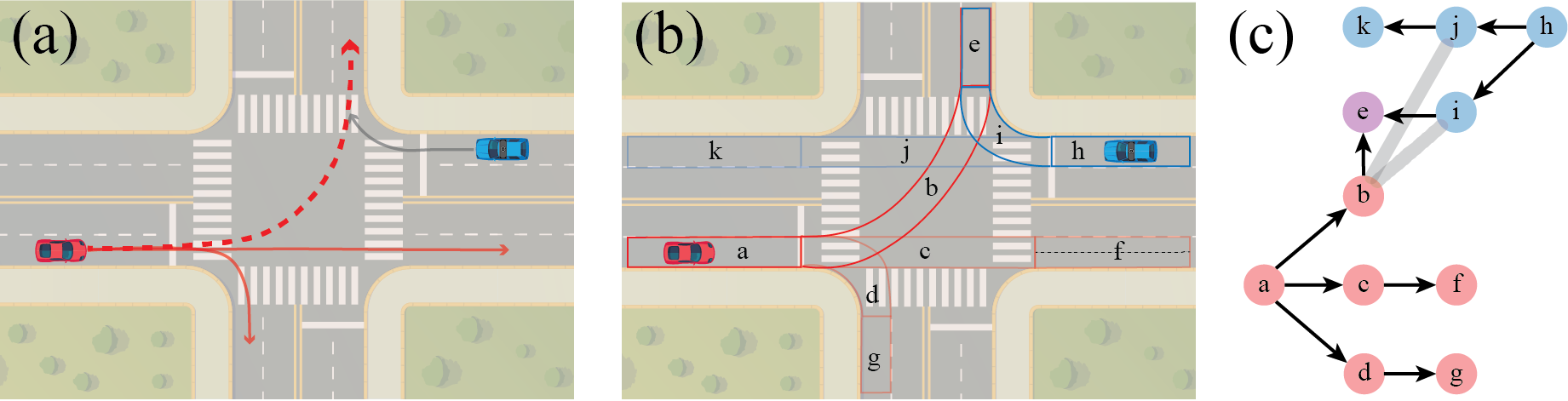}
\caption{While many feasible futures may exist for a \textcolor{red}{given actor}, only a small subset may be relevant to the \textcolor{blue}{AV's} planner. In \textbf{(a)}, neither of the dominant predicted modes (solid red) interact with the AV's intended trajectory (solid grey). Instead, the planner only needs to consider an illegal left turn across traffic (dashed red). \textbf{(b)} depicts a partial set of lane segments within the scene; illegal maneuvers such as following segment $b$ can either be mapped or hallucinated. A centerline (centered polyline) associated with a lane segment is shown in segment $f$ (dashed black). The planner can utilize the directed lane graph \textbf{(c)} to identify lanes which may interact with its intended route.  Black arrows denote directed edges, while thick grey undirected edges denote conflicting lanes.  Such networks are readily available in open street map APIs \cite{haklay2008openstreetmap} and the recently-released Argoverse \cite{Argoverse} dataset.}

\label{fig:multimodal}
\vskip -0.1in
\end{figure}

{\bf Explainability:} 
While the strong benchmark performance of feedforward models is encouraging, safety critical applications may require 
top-down feedback and causal explainability. For example, because the space of all potential futures in real-world urban driving settings is quite large, real-time planning may require the ability for a planner to interactively probe the forecaster, exploring only those futures that are relevant for planning (see Fig.\ref{fig:multimodal}a). Approaches that require re-generation or re-processing of the scene context in order to explore alternate futures may be too inefficient for real-time planning.

{\bf Our Approach:} In this paper, we develop a RNN-based approach for context-aware multi-modal behavior forecasting.  Our approach does not require rasterized input and includes both a road-network attention module and a dynamic interaction graph to capture interpretable geometric and social relationships.  In contrast to existing graph-based approaches \cite{gao2020vectornet, mercat2019multihead}, we structure our model to efficiently support counterfactual reasoning.  The social context of individual agents can be manipulated in order to condition upon additional hypothetical (unobserved) actors or to ablate specific social influences (Fig. \ref{fig:counterfactual_social}). We make intimate use of the road {\em network}, generating topological goals in the form of lane polylines that are constructed from the underlying directed graph of lane segments (Fig. \ref{fig:multimodal}c). Importantly, rather than encoding the full local map structure, we explicitly condition forecasts upon individual topological goals. This allows the planner to reason about and query for relevant trajectories (e.g. "reforecast that actor's motion given the left turn intersecting my path"). To our knowledge, we are the first to demonstrate counterfactual forecasts based on such topological queries.

\section{Related Work}

State-of-the-art models for multi-agent motion forecasting borrow heavily from both the natural language (sequence models) and computer vision (feature learning) communities. Although an extensive body of relevant work exists, the seq2seq \cite{sutskever2014sequence} and ResNet \cite{he2016deep} architectures are of particular theoretical and practical importance. Our work is most related to methods that forecast from intermediate representations such as tracking output \cite{mercat2019multihead}, although a significant body of work that operates directly on sensor input also exists \cite{Luo_2018_CVPR, casas2018intentnet}; we do not explore such approaches in-depth.


{\bf Motion Forecasting:} Until recently, motion forecasting research has primarily focused on pedestrian trajectories, either in the context of first-person activity recognition \cite{kitani2012activity}, sports \cite{zheng2016generating}, or multi-actor surveillance \cite{robicquet2016learning}. Social-LSTM \cite{alahi2016social} introduced social pooling within a RNN encoder-decoder architecture, providing a template to address varying numbers of actors and permutation problems caused by social input. Extensions such as \emph{SoPhie} \cite{Sadeghian_2019_CVPR} have leveraged features extracted from the physical environment, attention mechanisms, and adversarial training. DESIRE \cite{lee2017desire} proposed a scene-context fusion layer that aggregates interactions between agents and the scene context.  

{\bf Conditional Forecasting:} Recent work has also investigated forecasting models conditioned on intent: \cite{chai2019multipath, phan2019covernet} condition the agent's forecast on a predefined set of anchor trajectories, while \cite{deo2018multi, deo2018would} treat forecasting as a classification problem, first predicting a high level maneuver before conditioning predictions on that maneuver. Other methods, such as \cite{tang2019multiple}, predict a conditional probability density over the trajectories of other actors given a hypothetical rollout of the focal agent. Our work is most related to PRECOG \cite{Rhinehart_2019_ICCV}, which demonstrates that conditioning on an actor's goal alters the future states of other actors within the scene (similar to polyline conditioning).  Our approach, however, requires no rasterized input and can efficiently alter the social context as well.

{\bf Rasterization:} 
Popular methods for AV forecasting \cite{djuric2018motion, bansal2018chauffeurnet} have employed complex rasterized representations of scene context, constructing BEV images of the surrounding environment by combining trajectory histories with rich semantic information (lanes, speed limits, traffic light states, etc.) from maps. Although some of these methods \cite{djuric2018motion, cui2019multimodal, phan2019covernet} generate all predicted states simultaneously, others \cite{bansal2018chauffeurnet, chai2019multipath, tang2019multiple, Argoverse, mercat2019multihead, lee2017desire} employ a recurrent decoder to predict states sequentially; \cite{hong2019rules} experiments with both approaches. More recently, there has been interest towards rasterization-free approaches for capturing scene context; \cite{mercat2019multihead} uses multi-head attention to encode interactions between social actors. We adopt a similar approach, attending over lane polylines from the map, in addition to the social graph.



{\bf Graph Neural Networks:} Graph neural networks and graph convolution have emerged in response to problems that cannot be easily represented by matrices of pixels or simple vectors.  Architectures vary widely in response to diverse underlying problems and we refer the reader to \cite{wu2019comprehensive} and \cite{battaglia2018relational} for a comprehensive review. Our work is built upon graph attention networks (GATs), introduced in \cite{velivckovic2017graph}. VectorNet~\cite{gao2020vectornet} is a closely related method, which proposes a deep graphical model 
over individual road components and agent histories (represented as sets of vectors), claiming a 70\% reduction in model size compared to rasterized counterparts. Our work features similar advantages in parameter efficiency, but represents lane polylines as ordered sequences of points. Additionally, VectorNet conditions over the entire local neighborhood (e.g. left turn lane, right turn lane, neighboring lanes etc.) and is consequently not structured for counterfactual reasoning over specific map elements (e.g. right turn lane). Finally, VectorNet employs a deterministic decoder limited to a single trajectory.  In contrast, our approach employs a multi-modal decoder capable of generating diverse predictions.

\section{Method}
Our proposed architecture, the \textit{what-if} motion predictor (WIMP), addresses the task of motion forecasting by learning a continuous-space discrete-time system with $N$ interacting actors.
Let $\mathbf{x}^n_t \in \mathbb{R}^2$ denote the $n$-th actor's planar $(x,y)$ coordinates at time $t$ and $\mathbf{X}_t \doteq \left\{\mathbf{x}^1_t, \mathbf{x}^2_t, \dots, \mathbf{x}^N_t\right\}$ denote the joint state of all $N$ actors. Let $\mathbf{X} \doteq \left\{\mathbf{X}_1, \mathbf{X}_2, \dots, \mathbf{X}_t\right\}$ denote the joint observable history up until time $t$ and ${\bf X}^n = \{{\bf X}^n_1,{\bf X}^n_2, \ldots,{\bf X}^n_t\}$ represent the entire observable history for actor $n$. Analogously, let $\mathbf{Y} \doteq \left\{\mathbf{Y}_{t + 1}, \mathbf{Y}_{t+2}, \dots, \mathbf{Y}_{t+T}\right\}$ denote the joint state of all actors for future time-steps $t + 1$ to $t + T$. Let ${\bf Y}_t, {\bf Y}^n$, and ${\bf y}_t^n$ be defined accordingly.


{\bf Road Network Representation via Polylines:} 
Popular approaches for motion forecasting often rely on rasterized representations to provide contextual information about scene and road geometry \cite{bansal2018chauffeurnet, tang2019multiple, djuric2018motion}. Instead, we represent a valid path through the road network (directed graph of lane segments) using the concatenated center polylines of each road segment. Conditioning on polyline-based inputs has several advantages over its rasterized counterpart: i) it provides a strong, evidence-based prior for accurate predictions, ii) it allows for interpretable model behaviour analysis and enables counterfactual predictions that condition on hypothetical ``what-if" polylines (see Section \ref{sec:counterfactual}), and iii) it leads to more memory efficient models that do not require image-processing components.

\begin{figure}[t]
  \centering
  \begin{tabular}{cc}
  \includegraphics[width=0.5\textwidth]{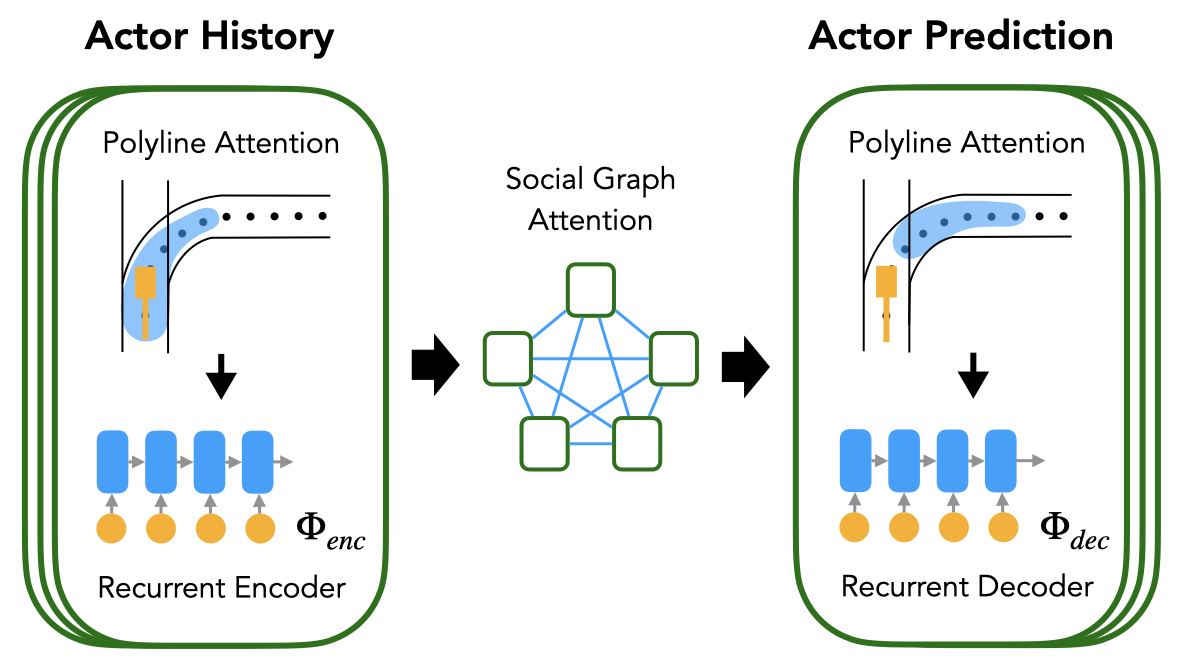} & \raisebox{0.02\height}{\includegraphics[width=0.45\textwidth]{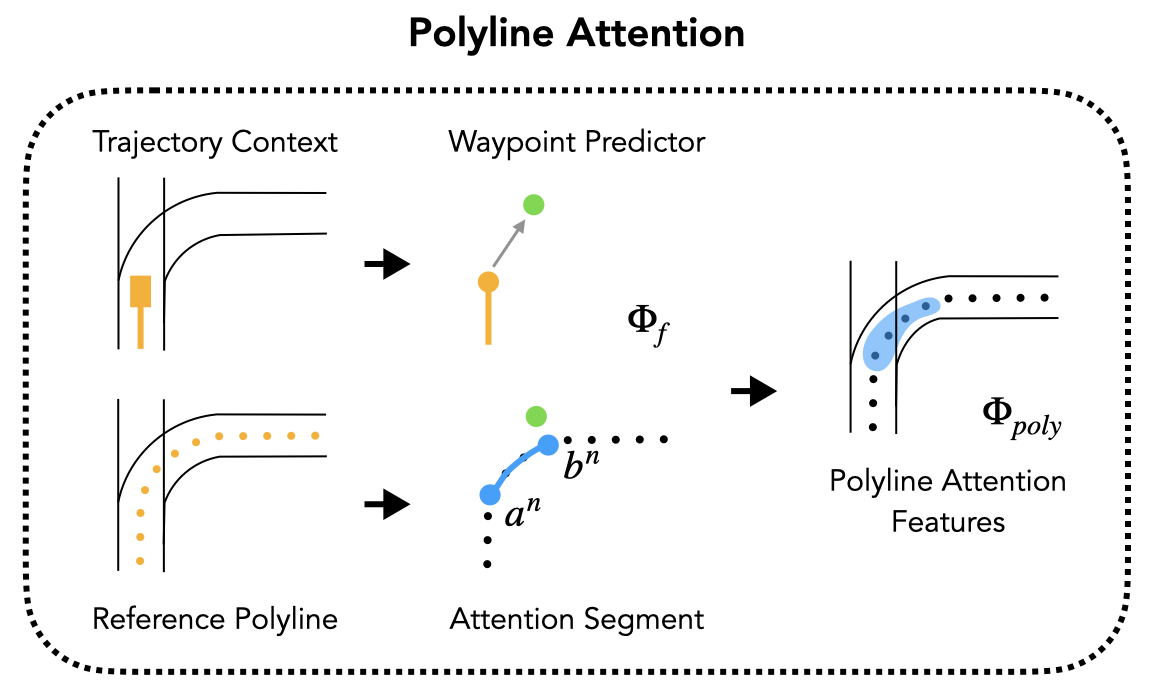}}
  \end{tabular}
  \caption{Overview of the data flow within the WIMP encoder-decoder architecture \textbf{(left)} and polyline attention module \textbf{(right)}. Input trajectories and reference polylines are first used to compute per-actor embeddings; social context is then incorporated via graph attention. Finally, a set of predictions is generated using a map-aware decoder that attends to relevant regions of the polyline via soft-attention.}
  \label{fig:architecture}
\end{figure}

We represent the reference polyline that guides actor $n$ as a set of $P$ discrete points $\mathbf{C}^n = \left\{\mathbf{c}^n_1, \mathbf{c}^n_2, \dots, \mathbf{c}^n_P \right\}$, where $\mathbf{c}^n_i \in \mathbb{R}^2$; the collective set of such polylines for all actors is denoted by $\mathbf{C} = \left\{\mathbf{C}^1, \mathbf{C}^2, \dots, \mathbf{C}^N \right\}$. Polyline $C^n$ is obtained by searching the road network along the direction of motion for the highest similarity lane segment to $X_n$ (additional details provided in Appendix \ref{apdx:polyline}). The final objective is to effectively model the conditional distribution $\Pr(\mathbf{Y}|\mathbf{X},\mathbf{C})$; though it is possible to model the aforementioned distribution in a joint fashion, it is often intractable and computationally inefficient for large $N$. Similar to \cite{tang2019multiple, mercat2019multihead}, we employ a RNN-based architecture to sequentially model $\Pr(\mathbf{Y}|\mathbf{X},\mathbf{C})$. Specifically, we assume that the following factorization holds:
\begin{equation}
    \label{eq:formulation}
    \Pr(\mathbf{Y}|\mathbf{X},\mathbf{C}) = \prod_{\delta=t+1}^{t + T} \Pr(\mathbf{Y}_\delta|\mathbf{Y}_{t + 1},\dots,\mathbf{Y}_{\delta - 1}, \mathbf{X}, \mathbf{C}) = \prod_{\delta=t+1}^{t + T}\prod_{n=1}^N \Pr(\mathbf{y}^n_\delta|\mathbf{Y}_{t + 1},\dots,\mathbf{Y}_{\delta - 1}, \mathbf{X}, \mathbf{C}^n)
\end{equation}
It should be noted that even though Eq.~\ref{eq:formulation} factorizes as a product of conditionals over individual actors conditioned on individual polylines, global information regarding other actors and polylines is \emph{implicitly} encapsulated via the history $\mathbf{X}$ and previous predictions $\left\{\mathbf{Y}_{t + 1},\dots, \mathbf{Y}_{\delta-1}\right\}$. To capture this distribution, we propose a novel recurrent, graph-based, attentional approach. As shown in Fig. \ref{fig:architecture}, the WIMP architecture has three key components: i) a graph-based encoder that captures scene context and higher-order social interactions, ii) a decoder that generates diverse, multi-modal predictions, and iii) a novel polyline attention mechanism that selects relevant regions of the road network to condition on. Next, we will describe each of these components in detail.

\subsection{Historical Context via Recurrence}
\label{sec:encoder}
\begin{equation}
\label{eq:encoder}
    \mathbf{h}_t^n = \Phi_{enc}\left(\mathbf{x}_t^n, \mathbf{s}_t^n, \mathbf{h}_{t-1}^n\right)\;\;,\;\;\mathbf{s}_t^n = \Phi_{poly}\left(\mathbf{C}^n, \mathbf{x}_t^n, \mathbf{h}_{t-1}^n\right)
\end{equation}
Each actor's contextual history $\mathbf{h}^t_n$ is captured via a shared recurrent encoder $\Phi_{enc}$. Similar to \cite{tang2019multiple}, we also employ a \emph{point-of-view} transformation $\Gamma(\mathbf{X^n})$ to normalize each actor's history to a reference frame by translation and rotation such that the $+x$-axis aligns with a focal agent $F$'s heading (such as the AV) and $\mathbf{x}^F_1 = (0,0)$.

\subsection{Geometric Context via Polyline Attention}
\label{sec:polyatt}
As described in Eq. \ref{eq:encoder}, each actor $n$ attends to segments of their reference polyline $\mathbf{C}^n$ through the learned function $\Phi_{poly}$. Intuitively, drivers pay attention to areas of the road network that they are {\em currently} close to, as well as future {\em goal} locations that they plan to reach. $\Phi_{poly}$ operationalizes this intuition by predicting, for each actor $n$ and timestep $t$, a current and goal index along its polyline:
\begin{equation}
    \label{eq:polysegment}
   a^n_t = \argmin_p\left\{d\left(\mathbf{c}^n_p, \mathbf{x}_t^n\right)\right\}\;\;,\;\; b^n_t = \argmin_p\left\{d\left(\mathbf{c}^n_p, \Phi_{f}\left(\mathbf{x}_t^n, \mathbf{h}_{t-1}^n, \Delta \right)\right)\right\}
\end{equation}
where $d(\cdot)$ is a distance metric and $\Phi_f$ is a learned function that hallucinates a coarse {\em waypoint} $\Delta$ time-steps in the future. It should be noted that $\Phi_f$ doesn't make use of any polyline information and predicts the waypoint solely based on kinematic history; training is conducted in a self-supervised manner using ground-truth future trajectories as labels. The vectorized attention-weighted representation $\mathbf{s}^n_t$ for the segment $\mathbf{\bar{C}}^n_t$ between current and goal indices can then be obtained as follows (where $\mathbf{Q}, \mathbf{V}, \mathbf{K}$ are learned transformation matrices, similar to those employed in \cite{mercat2019multihead}):
\begin{equation}
    \label{eq:polyatt}
   \Phi_{poly}({\bf C}^n,{\bf x}_t^n,{\bf h}_{t-1}^n) = \sum_{r\in [a^n_t,b^n_t]}\upsilon^n_{tr}\mathbf{V}\mathbf{c}^n_r\;\;,\;\;\upsilon^n_{tr} = \underset{r}{\text{softmax}}\left(\mathbf{Q}\mathbf{h}_{t-1}^n \odot \mathbf{K}\mathbf{c}^n_r\right)
\end{equation}

\subsection{Social Context via Graph Attention}
As $\Phi_{enc}$ runs independently over all actors, the hidden representation obtained after $t$ time-steps $\mathbf{h}_t^n$ for a particular actor $n$ is oblivious to other dynamic participants in the scene. One possible solution is to provide $\mathbf{x}_t^i; \forall i\neq n$ as an input to Eq. \ref{eq:encoder}, but this is computationally inefficient and memory intensive. Instead of capturing social interactions in the planar coordinate space, we leverage the ability of $\Phi_{enc}$ to generate rich latent hidden representations $\mathbf{h}_t^n$ for a particular actor $n$. Inspired by \cite{velivckovic2017graph}, we employ a graph attention module $\Phi_{gat}$ that operates over these representations as follows:
\begin{equation}
    \label{eq:gat}
    \mathbf{\bar{h}}_{t}^n = \sigma\left(\mathbf{h}_t^n + \frac{1}{D}\sum_{d=1}^{D}\sum_{j\in N\setminus n} \alpha_{nj}^d\mathbf{W}^d\mathbf{h}_t^j\right)\; , \; \alpha_{nj}^d = \underset{j}{\text{softmax}}\left(\mathbf{a}^{d}\odot\left[\mathbf{W}^d\mathbf{h}_t^n, \mathbf{W}^d\mathbf{h}_t^j\right]\right)
\end{equation}
where $D$ is a hyperparameter denoting the number of attention heads, $[\cdot,\cdot]$ is the concatenation operation, $\odot$ is the inner product, and $\mathbf{W}^d$, $\mathbf{a}^d$ are learned parameters. Note that there is a subtle difference between Eq. \ref{eq:gat} and the architecture proposed in \cite{velivckovic2017graph}, wherein, for each agent $n$, we focus on learning a residual change to its \emph{socially-unaware} hidden representation $\mathbf{h}_t^n$. Intuitively, this can be thought of as an actor initially having a socially-agnostic estimate of its future trajectory, with $\Phi_{enc}$ learning a residual change to incorporate information from other actors within the scene.

\subsection{Decoding}
\label{sec:decoder}
Following Eq. \ref{eq:formulation}, WIMP aims to learn the conditional distribution $\Pr(\mathbf{y}^n_\delta|\mathbf{Y}_{t + 1},\dots,\mathbf{Y}_{\delta - 1}, \mathbf{X}, \mathbf{C}^n)$ for each actor $n$. To achieve this goal, we employ a LSTM-based decoder $\Phi_{dec}$ that: i) generates diverse and multi-modal predictions, and ii) conditions each prediction on a reference polyline $\mathbf{C}^n$. Particularly, for a future time-step $\delta$, we can obtain $\mathbf{y}^n_\delta$ as follows:
\begin{equation}
    \label{eq:decoder}
    \mathbf{y}^n_{\delta+1} = \Phi_{pred}\left(\mathbf{o}^n_{\delta}\right)\;\;,\;\;\mathbf{o}^n_\delta, \mathbf{\bar{h}}_{\delta}^n = \Phi_{dec}\left(\mathbf{Y}_\delta, \mathbf{\bar{s}}_\delta^n, \mathbf{\bar{h}}_{\delta-1}^n\right)\;\;,\;\;\mathbf{\bar{s}}_\delta^n = \Phi_{poly}\left(\mathbf{C}^n, \mathbf{y}_\delta^n, \mathbf{\bar{h}}_{\delta-1}^n\right) 
\end{equation}

where $\Phi_{pred}$ is a learned prediction function and $\Phi_{poly}$ is a polyline-attention module as described in Section \ref{sec:polyatt}. We note that the implementation of $\Phi_{pred}$ is architecturally agnostic; for example, $\Phi_{pred}$ could be a bivariate Gaussian as in \cite{tang2019multiple}, or a mixture of Gaussians as in \cite{mercat2019multihead}. For datasets like Argoverse~\cite{Argoverse} that only evaluate predictions for a single focal actor $F$, decoder input $\mathbf{Y_\delta}$ might only contain predictions for a single actor $\mathbf{y}^F_{\delta}$. However, even in this scenario, WIMP is still able to model social interactions via embeddings $\mathbf{\bar{h}}_{t}^n$ obtained from the graph-based encoder. 

\begin{figure}[t!]
  \centering
  \includegraphics[width=\linewidth]{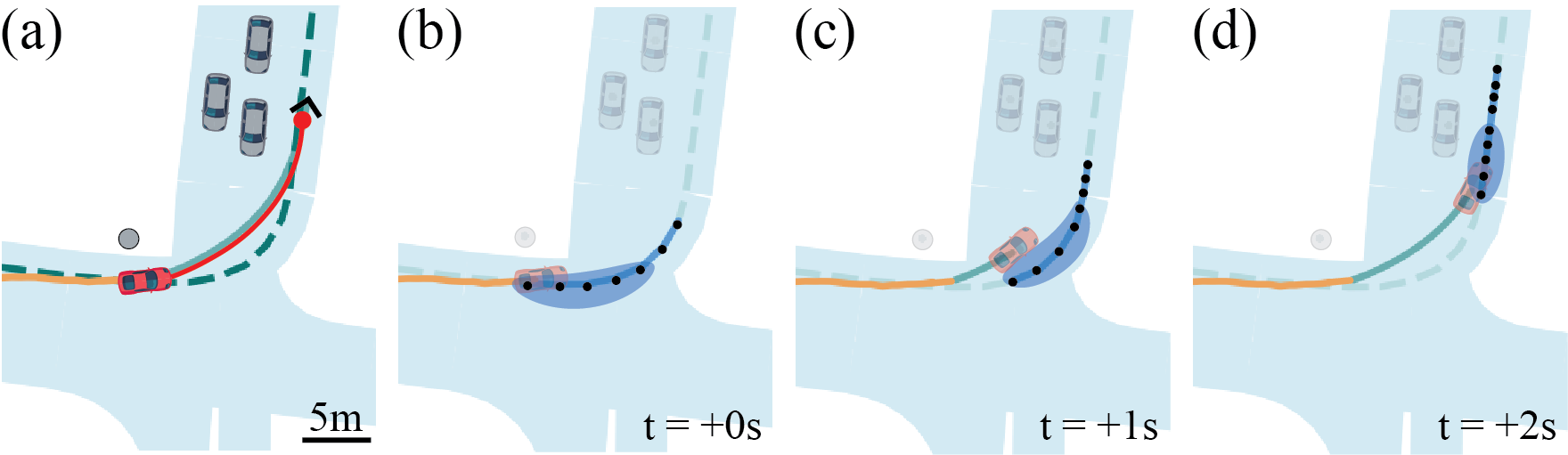}
  \caption{Visualizing the \textit{map lane polyline} attention weights generated during decoding. In the scenario depicted in \textbf{(a)}, the \textcolor{red}{focal actor's} history is shown in yellow and its ground-truth future in red.  The red circle highlights the true state 3s into the future.  The solid green line denotes a predicted trajectory with a black chevron marking the $t=+3s$ state. The dashed green line shows the reference polyline.  Grey cars/circles illustrate the current positions of on/off roadway actors. In \textbf{(b, c, d)}, opacity corresponds to the magnitude of social attention. The subset of the polyline selected by the polyline attention module is shown in solid blue (points denoted as black circles), and the attention weights within that segment are shown via an ellipse (for predictions at $t=+0s, +1s, +2s$ respectively).  Points outside the ellipse have negligible attention. WIMP learns to attend smoothly to upcoming points along the reference polyline.}
  \label{fig:vizattention}
\end{figure}

\subsection{Learning}
\label{sec:learning}
WIMP is trained on collections of triplets containing: historical trajectories, ground-truth future trajectories, and map-based road context $\{({\bf X},{\bf Y},{\bf C})\}$. Following standard forecasting benchmarks, we only predict the future trajectory for a single {\em focal} agent in each training example, denoted as ${\bf Y}^F$.

{\bf Winner-Takes-All:} To encourage diversity and multi-modality in the set of predicted trajectories, we learn a mixture of $M$ different predictors. Diversity is encouraged through a ``multiple choice" \cite{guzman2012multiple} or ``winner-takes-all" loss that explicitly {\em assigns} each training example to a particular mixture:
\begin{align}
\text{loss} = \min_{m \in \{1 \ldots M\}} ||{\bf \hat Y}^F_m - {\bf Y}^F||
\end{align}
where ${\bf \hat Y}^F_m$ is the focal trajectory predicted by the $m^{th}$ mixture. Having experimented with various distance functions, we found the $L1$ norm between trajectories to perform well. We also experimented with {\em multi-agent} prediction of future trajectories for {\em all} actors, but did not observe improved performance. We posit that this may be due to the large numbers of parked actors present in urban driving scenarios, which may require different representations or larger capacity forecasting models.

{\bf Optimization:} By keeping track of the $\argmin$ $m$ index for each training example, WTA loss naturally clusters training examples into $M$ sets. Previous work has shown that directly optimizing this loss can lead to poor results because (a) it is difficult to optimize stochastically with mini-batch SGD, as the optimization is sensitive to initialization and (b) each mixture can be prone to overfitting, as it is trained with less data. One proposed solution is ``evolving WTA" (EWTA) \cite{makansi2019overcoming}, where the single minimum $\min_m$ is replaced with the $M'$ lowest-cost mixtures. Initializing with $M'=M$, examples are initially associated with all $M$ clusters, encouraging every mixture to generate identical predictions. Over time, as $M'$ is annealed to 1 (resulting in standard WTA loss), iterative specialization of each mixture ensures that each of the final mixtures has been ``pre-trained" with the full
dataset.

{\bf Mixture Ranking:} The above produces $M$ different predicted trajectories, which can be fed directly into multi-output forecasting benchmarks that require methods to return $M$ predictions. To repurpose these outputs for single-prediction evaluations, we rank each mixture's accuracy on a validation set. 

\section{Experiments}
We demonstrate the effectiveness of WIMP at generating accurate, interpretable, and controllable trajectory predictions for roadway actors. We first show that the scene attention encoder is capable of capturing the complex contextual, semantic, and social relationships that are present in real-world urban driving scenarios. These learned scene embeddings can be combined with multi-modal decoders to generate a diverse set of plausible future trajectories. We then perform a series of counterfactual reasoning-based experiments to demonstrate how the distribution of predicted modes is influenced by scene context. The implementation details and hyper-parameters are provided in Appendix \ref{apdx:details}.

\subsection{Experimental Setup}
\textbf{Datasets.}
\label{sec:exp_datasets}
We conduct our experiments using the Argoverse \cite{Argoverse} motion forecasting dataset, a large scale vehicle trajectory dataset containing more than 300,000 curated scenarios extracted from vehicle logs in urban driving scenarios. Given a 2 second trajectory history as input, the goal is to predict the future motion of a particular focal agent over the next 3 seconds (sampled at $\approx$ 100 ms intervals). In addition to the focal agent history, location histories of nearby (social) actors are also provided.  Importantly, Argoverse includes a semantic vector map composed of lane-based polylines.

Although the Argoverse dataset provides a high volume of \textit{interesting} data for both training and evaluation, the focal trajectories are not particularly diverse in terms of directional variation, with more than 80\% of scenarios featuring straight line trajectories over the full 5 second window. In order to evaluate how WIMP performs in the presence of uncertainty, we also extract a small subset ($\approx$350 examples) of particularly challenging scenarios that are characterized by blind turns (defined as examples where the observed 2-sec. trajectory is straight, but the ground truth future 3-sec. trajectory contains a turn and/or lane change). Even for recent state-of-the-art methods, the blind turn (BT) subset presents a significant challenge, as generation of high-quality predictions necessitates the incorporation of both social and semantic information to resolve uncertainty.

In addition to Argoverse, we also evaluate using the NuScenes prediction dataset, which contains a similar collection of approximately 40,000 scenarios that were extracted from 1,000 curated scenes. These scenarios were collected within two distinct regions on different continents, featuring trajectories from both left-hand and right-hand drive locales. Due to the geographic diversity of data and more significant representation of complex scenarios such as turns and intersections, NuScenes presents a more challenging prediction task than the base Argoverse dataset.

\textbf{Metrics.}
To evaluate prediction quality, we make use of widely adopted forecasting metrics: minimum average displacement error (ADE) and minimum final displacement error (FDE) \cite{Argoverse}, evaluated for both single ($K = 1$) and multi-modal $(K = 6)$ prediction scenarios. To capture prediction performance in more challenging scenarios, we also adopt the miss rate (MR) metric: the fraction of scenarios with FDE $> 2m$.

\subsection{Quantitative Results}
\textbf{Argoverse Motion Forecasting.} We compare WIMP to several recent state-of-the art (SOTA) methods: SAMMP \cite{mercat2019multihead} (self-attention-based model, joint-winner of the 2019 Argoverse Forecasting Challenge), UULM-MRM (rasterization-based model, joint-winner of the 2019 Argoverse Forecasting Challenge), VectorNet \cite{gao2020vectornet} (recent polyline-based model), and LaneGCN \cite{liang2020learning} (concurrently-developed lane graph-based model). 
Evaluating on the Argoverse challenge test set (results summarized in Table \ref{table:testresults}), we show that each of these methods is highly competitive, performing far above the bar set by K-NN and LSTM based baselines. We further show that WIMP out-performs all prior published work and achieves similar performance to concurrent work, while providing unique advantages in flexibility. 
Lastly, because many of the top-ranked entries in the Argoverse challenge do not provide descriptions of their methodology, we compare against such methods in Appendix \ref{apdx:challengeresults}.

\begin{table*}[h]
\begin{center}
\begin{small}
\begin{sc}
\begin{tabular}{lccccc}
\toprule
Model & MR(K=6) & FDE(K=6) & ADE(K=6) & FDE(K=1) & ADE(K=1)\\
\midrule
LaneGCN \cite{liang2020learning}  & \textbf{0.16} & \textbf{1.36} & \textbf{0.87} & \textbf{3.78} & \textbf{ 1.71}\\
SAMMP \cite{mercat2019multihead}  & 0.19 & 1.55 & 0.95 & 4.08 & 1.81\\
UULM-MRM & 0.22 & 1.55 & 0.96 & 4.32 & 1.97\\
NN + Map(Prune) \cite{Argoverse} & 0.52 & 3.19 & 1.68 & 7.62 & 3.38\\
LSTM + Map(Prior) \cite{Argoverse} & 0.67 & 4.19 & 2.08 & 6.45 & 2.92 \\
VectorNet\cite{gao2020vectornet} & - & - & - & 4.01  & 1.81\\
\midrule
WIMP ($M=1$) & - & - & - & 3.89 & 1.78\\
WIMP ($M=6$) & 0.17 & 1.42 & 0.90 & 4.03 & 1.82  \\
\bottomrule
\end{tabular}
\end{sc}
\end{small}
\end{center}
\caption{Motion forecasting performance evaluated on the Argoverse test set, with MR and minimum FDE/ADE reported for both single ($K=1$) and multi-modal ($K=6$) prediction scenarios.}
\label{table:testresults}
\end{table*}

\begin{wraptable}{r}{0.5\textwidth}
\vspace{-7mm}
\centering
\begin{tabular}{lccc}\\
\toprule
Model & MR & FDE & ADE\\
\midrule
SAMMP  & 0.67 & 4.91 & 2.38\\
NN + Map (Prune) & 0.61 & 5.11 & 3.93\\
LSTM + Map (Prior) & 0.51 & 2.64 & 3.01 \\
\midrule
WIMP & 0.49 & 3.52 & 1.62\\
WIMP (Oracle) & \textbf{0.33} & \textbf{2.46} & \textbf{1.30}\\
\bottomrule
\end{tabular}
\caption{Motion forecasting performance evaluated on the Argoverse BT validation set. As the selected data is inherently multi-modal, we only report metrics for ($K=6$) predictions. SAMMP results were obtained from our implementation of \cite{mercat2019multihead}, using hyper-parameters shared with WIMP.}
\vspace{-3mm}
\label{table:btresults}
\end{wraptable}

\textbf{Evaluation in Challenging Scenarios.} As the overall Argoverse dataset is biased towards simple straight line trajectories, we also evaluate prediction performance on the BT subset (results summarized in Table \ref{table:btresults}), which consists primarily of challenging blind turn scenarios. In this setting, we show that WIMP out-performs non-map-based approaches (such as SAMMP) by a much larger margin than across the full dataset, as polyline and social graph-based attention allows the model to resolve and account for uncertainty even in complex scenarios with multiple feasible future trajectories. In such scenarios, models employing polyline-based coordinate systems, such as LSTM + Map (Prior) from \cite{Argoverse}), also perform surprisingly well, as the prediction space is strongly conditioned on map information, trading overall performance for better turn prediction results. We note that WIMP is significantly less impacted by this bias-variance trade-off, delivering top performance in both BT and general settings. We also demonstrate that prediction accuracy improves with reference polyline quality. By employing an oracle to select the optimal polyline in hindsight (after observing the future), we observe significant improvements, indicating that WIMP can take advantage of ``what-if" polylines provided by such oracles. We analyze this further in the next section.


\textbf{Ablation Study}
In order to demonstrate how each component of the WIMP architecture contributes to overall prediction performance, we perform an ablation study and summarize the results in Table \ref{table:ablation}. We obtain best results when the model is provided with both map and social context, while coupled to a L1-based EWTA loss \cite{makansi2019overcoming}. We also experiment with alternative loss formulations: replacing EWTA loss with negative log likelihood (NLL) significantly degrades performance, while standard L1 loss provides impressive ($K=1$) performance but cannot be adapted to make multiple predictions.

\begin{table*}[h]
\label{ablationtable}
\begin{center}
\begin{small}
\begin{sc}
\begin{tabular}{ll|ccccc}
\toprule
Context & Loss & MR(K=6) & FDE(K=6) & ADE(K=6) & FDE(K=1) & ADE(K=1)\\
\midrule
Map + Social & EWTA & \textbf{0.12} & \textbf{1.14} & \textbf{0.75} & 3.19 & 1.45  \\
Map + Social & L1 & - & - & - & \textbf{3.01} & \textbf{1.40}\\
\midrule
Map + Social & NLL & 0.23 & 1.61 & 1.07 & 6.37 & 1.41\\
Social & EWTA & 0.16 & 1.39 & 0.86 & 5.05 & 1.61\\
Map & EWTA & 0.16 & 1.38 & 0.85 & 3.80 & 1.69\\
None & EWTA & 0.23 & 1.70 & 0.95 & 5.86 & 1.87\\
\bottomrule
\end{tabular}
\end{sc}
\end{small}
\end{center}
\caption{Ablation studies for WIMP with different input configurations and training objectives. Quantitative results reported for ($K=1$) and ($K=6$) metrics on the Argoverse validation set.}
\label{table:ablation}
\vspace{-1mm}
\end{table*}

\textbf{NuScenes Motion Forecasting.} To evaluate the generalizability of our proposed prediction architecture, we also compare WIMP to several recent learning-based methods and a physics-based baseline on the NuScenes prediction dataset; results are summarized in Table \ref{table:nuscenes}. Without any hyper-parameter tuning or changes to model architecture (compared to the model evaluated on Argoverse), WIMP achieves state-of-the-art results in both single ($K=1$) and multi-modal ($K=5, 10$) prediction scenarios, out-performing all previous methods in both miss rate and displacement error. WIMP delivers especially strong results on NuScenes due to the high proportion of intersection and turn-based scenarios, which are difficult to solve without integration of both map and social context.

\begin{table*}[h]
\begin{center}
\begin{small}
\begin{sc}
\setlength\tabcolsep{1.5pt}
\begin{tabular}{lcccccc}
\toprule
Model & MR(K=10) & ADE(K=10) & MR(K=5) & ADE(K=5) & FDE(K=1) & OffRoadRate\\
\midrule
LISA & 0.46 & 1.24 & 0.59 & 1.81 & 8.57 & 0.07\\
Trajectron++\cite{salzmann2020trajectron++} & 0.57 & 1.51 & 0.70 & 1.88 & 9.52 & 0.25\\
cxx  & 0.60 & 1.29 & 0.69 & \textbf{1.63} & 8.86 & 0.08\\
CoverNet\cite{phan2019covernet} & 0.64 & 1.92 & 0.76 & 2.62 & 11.36 & 0.13\\
Physics Oracle\cite{phan2019covernet} & 0.88 & 3.70 & 0.88 & 3.70 & 9.09 & 0.12 \\
\midrule
WIMP & \textbf{0.43} & \textbf{1.11} & \textbf{0.55} & 1.84 & \textbf{8.49} & \textbf{0.04} \\
\bottomrule
\end{tabular}
\end{sc}
\end{small}
\end{center}
\caption{Motion forecasting performance evaluated on the NuScenes validation set, with MR and minimum FDE/ADE reported for ($K=1$), ($K=5$), and ($K=10$) prediction scenarios. Metrics are computed using the reference implementation provided with the NuScenes devkit.}
\label{table:nuscenes}
\vspace{-0.2cm}
\end{table*}

\subsection{Counterfactual Validation}
\label{sec:counterfactual}



\begin{figure}[t]
 \centering
\includegraphics[width=.9\linewidth]{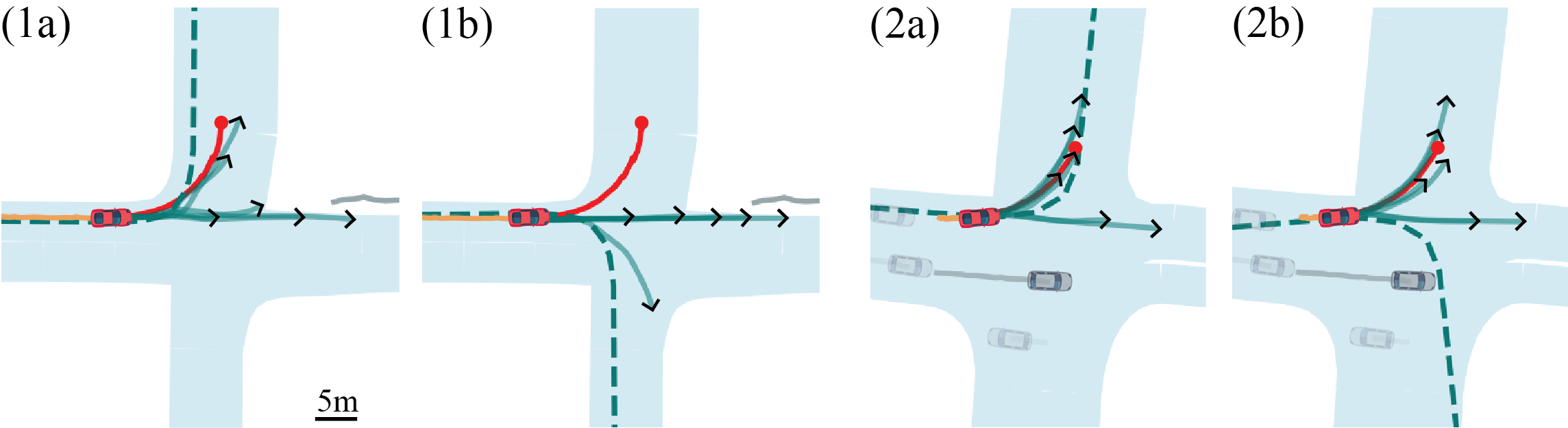}
\caption{Visualizations of two prediction scenarios that condition on \textbf{(a)} heuristically-selected polylines (see Appendix \ref{apdx:polyline} for details) and corresponding \textbf{(b)} counterfactual reference polylines. When making diverse predictions, WIMP learns to generate some trajectories independent of the conditioning polyline (see the straight through predictions in \textbf{(a)}). Additionally, if the reference polyline is semantically or geometrically incompatible with the observed scene history (as in \textbf{(2b)} where the counterfactual polyline intersects other actors), the model learns to ignore the map input, relying only on social and historical context. Visualization style follows Fig. \ref{fig:vizattention}.
}
\label{fig:counterfactual_cl}
\end{figure}

\begin{figure}[t]
  \centering
  \includegraphics[width=.85\linewidth]{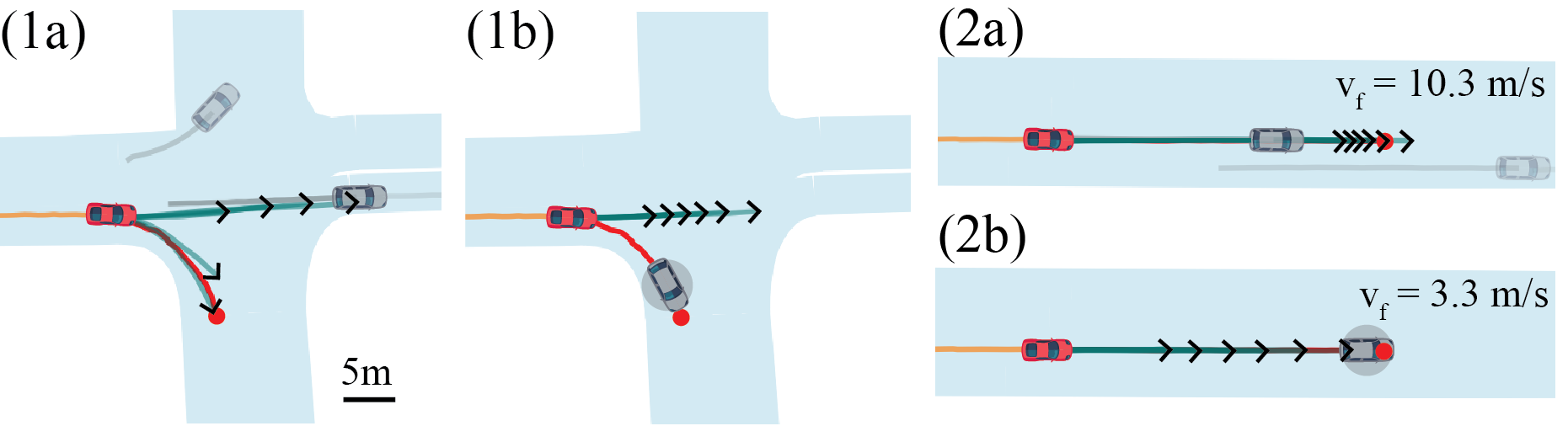}
  \caption{Visualizations of two scenarios that condition on \textbf{(a)} ground-truth scene context and \textbf{(b)} counterfactual social contexts (best viewed with magnification).  Counterfactual actors are highlighted with a grey circle. In \textbf{(1b)}, we inject a stopped vehicle just beyond the intersection, blocking the ground-truth right turn.  Given the focal agent's history and velocity, this makes a right turn extremely unlikely, and that mode vanishes. In \textbf{(2b)} we replace the the leading actor in \textbf{(2a)} with a stopped vehicle.  As expected, this causes the model to predict trajectories containing aggressive deceleration.  The final velocity ($v_f$) of a representative trajectory is 3.3m/s in the counterfactual setting, compared with 10.3m/s in the original scene. Visualization style follows Fig. \ref{fig:vizattention}.}
  \label{fig:counterfactual_social}
\vspace{-2mm}
\end{figure}

Our proposed approach to conditional forecasting readily supports investigations of hypothetical or unlikely scenarios (counterfactuals).  This capability can be readily used by a planner to allocate computation to only relevant futures, or to reason about social influences from occluded regions of the road network.  Importantly, these counterfactual queries can also be used to investigate and evaluate models beyond distance-based metrics.  Sensible predictions conditioned on extreme contextual input indicates that our model has learned a powerful causal representation of driving behavior and is likely to generalize well (see Figs. \ref{fig:counterfactual_cl} and \ref{fig:counterfactual_social}).   


\section{Discussion}
In this paper, we proposed a recurrent graph-based attentional framework with interpretable geometric and social relationships that supports the injection of counterfactual contextual states. Leveraging information from historical, social, and geometric sources, WIMP facilitates joint multi-modal prediction of future states over an arbitrary number of actors within a scene, out-performing all previous methods on the Argoverse forecasting dataset. In future work, it would be interesting to extend the polyline selection procedure with an end-to-end trainable solution, enabling the model to automatically select candidate polylines based on observed scene context. Alternative directions for future research could explore applications of WIMP beyond autonomous driving, perhaps for prediction of pedestrian trajectories or human actions; we will release code\footnote{Code will be released at \href{https://github.com/wqi/WIMP}{https://github.com/wqi/WIMP}.} to facilitate such work.

\section*{Broader Impact}
The ability to plan, conditioned on the future states of dynamic agents in complex roadway environments is a central challenge to the safe and efficient operation of autonomous vehicles.  Progress on the motion prediction problem has downstream consequences for the deployment timeline, scale, and performance of autonomous vehicles as paratransit, long-haul freight, and local delivery options.  Implications of the mass deployment of AVs are examined and simulated in an increasing number of economic \cite{clements2017economic}, public policy \cite{anderson2014autonomous, litman2017autonomous}, and most recently public health \cite{crayton2017autonomous} papers.  We refer the reader to \cite{milakis2017policy} and \cite{anderson2014autonomous} for a literature review and holistic synthesis respectively.


Independent of the implications of autonomous vehicles deployment at scale, this project is an explicit attempt to impact and focus future research on motion forecasting in this domain.  Recent work has proposed innovative architectures and achieved impressive benchmark performance, however often without consideration for the pragmatic requirements necessary to deploy these architectures on public roadways.  (1) Prediction systems serve the planner.  Proposed prediction models should discuss how predictions can be utilized by the planner to address difficult scenarios safely. (2) Learned AV systems must respond well to novel scenarios outside the domain of training data.  Models should demonstrate prudent behavior in response to extreme scenarios or perturbed inputs. (3) AV subsystems must be interpretable.  Interpretable and traceable decisions are necessary to build compelling safety cases for both regulatory approval and public trust.

\begin{ack}
This work was supported by the CMU Argo AI Center for Autonomous Vehicle Research.
\end{ack}

{\small
\bibliographystyle{plain}
\bibliography{ref}
}
\clearpage

{\LARGE \textbf{Appendix}}
\appendix

\section{Implementation Details}
\label{apdx:details}
Although we demonstrate the WIMP forecasting framework using a vehicle trajectory prediction task in an autonomous driving setting, the architecture is designed such that concrete implementations of learned components and transformations are abstracted away. This improves generalization, as a variety of prediction tasks can be supported through the selection of specific component configurations. To improve reproducibility, we share the specific implementation details (including hyper-parameters for training and model configuration) for the WIMP model we use to report results on Argoverse. 

\textbf{Normalization.} Prior to all other operations, every collection of points specified in global Argoverse world coordinates (input trajectories, reference polylines, etc.) within each scenario is first transformed to a local coordinate space that is normalized with respect to focal agent $F$. This is implemented using an affine transformation $A$, such that the positive X-axis becomes aligned with the focal agent's heading (defined as the angle between $\mathbf{x}^F_1$ and $\mathbf{x}^F_{20}$) and $\mathbf{x}^F_1 = (0,0)$

\textbf{Polyline Attention.}
Both the encoder $\Phi_{enc}$ and decoder $\Phi_{dec}$ make use of polyline attention module $\Phi_{poly}$ to capture priors provided by the map. However, weights are not shared between the two polyline attention modules. This is largely a consequence of $\Phi_{dec}$ only predicting a trajectory for the focal agent $F$ (owing to the task formulated by the Argoverse \cite{Argoverse} dataset), whereas $\Phi_{enc}$ takes observed trajectories from all actors as input. $\Phi_{poly}$ is implemented as a $4$-layer LSTM, where the hidden state is a $4\times512$ dimensional vector. The distance metric $d(\cdot)$ in Equation 3 is the L2-norm. The transformations $\mathbf{Q}, \mathbf{K}, \mathbf{V}$ defined in Equation 4 are learned matrices of size $512\times512$ and are used in the same manner as \cite{mercat2019multihead, vaswani2017attention}. We use a dropout \cite{srivastava2014dropout} rate of 0.5 during training, which is applied over the first three layers.

\textbf{Encoder.} The shared recurrent encoder $\Phi_{enc}$ used to capture each actor's location history is implemented as a 4-layer LSTM with a 512-dimensional hidden state. We use a dropout rate of 0.5 during training, which is applied over the first three layers.

\textbf{Graph Attention.} The graph attention module $\Phi_{gat}$ takes as input the final hidden state $\mathbf{h}^t_n$, for each actor $n$. Following Equation 5 of the main paper, we set the number of attention heads $D$ to 4, and the learned parameters $\mathbf{W}^d$ and $\mathbf{a}^d$ are of sizes $2048 \times 512$ and $1024 \times 1$ respectively.

\textbf{Decoder.} The decoder $\Phi_{dec}$ is configured identically to the encoder $\Phi_{enc}$, wherein we use a 4-layer LSTM with a 512-dimensional hidden state. Following Equation 6 of the main paper, $\Phi_{pred}$ is a linear layer that transforms the $512$-dimensional output $\mathbf{o}_{\delta}^n$ of $\Phi_{dec}$ into a $2$-dimensional prediction.

\textbf{Training.}
For training on the Argoverse dataset, we use the ADAM optimizer with stochastic mini-batches containing 100 scenarios each; no weight decay is employed, but gradients are clipped to a maximum magnitude of 1.0. The learning rate is initialized to a value of 0.0001 and annealed by a factor of 2 every 30 epochs.  We couple the optimizer to an EWTA-based loss (as described in Section 3.5), the value of $M'$ is intialized to 6 and annealed by 1 every 10 epochs until $M' = 1$ at epoch 50. Validation metrics are computed after every 3 training epochs and training is terminated once validation metrics have failed to improve for 30 epochs in a row. Each model requires approximately 100 epochs to train on average, taking about 28 hours of compute time on an AWS ``p3.8xlarge" instance equipped with 4x V100 GPUs.

\textbf{Evaluation.} As predictions are generated in the normalized local coordinate space, they are first transformed back to the global world space using an inverse affine transformation $A^{-1}$ before evaluation. The minimum final displacement error (minFDE) metric is computed by taking the minimum of L2 distances between the end points of each of the $k$ predicted trajectories and the ground truth future; minimum average displacement error (minADE) is then obtained by computing the average L2 distance corresponding to the predicted trajectory with lowest end point error.  Finally, we compute the miss rate, which measures the proportion of scenarios where minFDE exceeds a threshold value (set at 2m in the Argoverse Forecasting Challenge).

\section{Polyline Selection}
\label{apdx:polyline}
To obtain relevant reference trajectories from the underlying vector map, we employ a heuristic-based polyline proposal module based on code released in the Argoverse API \cite{Argoverse}. Using either the observed (0-2s) history of the focal actor (during evaluation) or the full (0-5s) ground truth trajectory (during training), we query the proposal module for a ranked list of candidate polylines sorted by similarity to the reference trajectory. These candidate polylines are obtained through the following procedure:
\begin{enumerate}
    \item \textbf{Find Candidate Lanes}: We first search the map lane graph to find the set of all lanes containing nodes that are located within a 2.5m distance from the last point of the query trajectory. If no lanes are found, we iteratively expand the search radius by a factor of 2 until at least one candidate lane is identified.

    \item \textbf{Construct Candidate Polylines}: For each candidate lane node returned in the above set, we construct corresponding polylines by recursively traversing the lane graph through successor and predecessor nodes, stopping once a distance threshold has been reached in both directions. In our implementation, we set this distance threshold to be $2\times$ the total length of the query trajectory. We then connect the traversed nodes with directed edges (from earliest predecessor to latest successor), forming a polyline composed of individual points. To show how candidate polylines $L_c$ are constructed from candidate lane node $A$:

    \begin{center}
    Enumerated Successors: \textit{\{(A->B->C), (A->D->E)\}} \\
    Enumerated Predecessors: \textit{\{(F->G->A), (H->I->A)\}} \\ 
    Constructed Lane Polylines: \\
    \textit{L1 : F->G->A->B->C} \\
    \textit{L2 : H->I->A->B->C} \\
    \textit{L3 : F->G->A->D->E} \\
    \textit{L4 : H->I->A->D->E}  \\
    \textit{$L_{c}$: \{L1, L2, L3, L4\}}
    \end{center}

    \item \textbf{Remove Overlapping Polylines}: Next, we filter the set of candidate polylines constructed in the previous step by removing polylines that overlap significantly with other candidates.

    \item \textbf{Sort By Point-in-Polygon Score}: We then sort the filtered set of candidate polylines by point-in-polyline (PIP) score, defined as the number of query trajectory points that lie within the polygon formed by lane regions corresponding to each polyline. If there are $n$ points in the query trajectory, the PIP score is bounded to the range $[0, n]$. To give a concrete example, if $n=20$ and PIP scores for candidate polylines $L_{c}$ are \textit{\{L1: 15, L2: 10, L3: 5, L4: 20\}}, the sorted list of candidate polylines will be returned in the order $L_{pip} = \textit{[L4, L1, L2, L3]}$.

    \item \textbf{Sort By Polyline-Trajectory Alignment}: We also sort the filtered set of candidate lines by a polyline-trajectory alignment-based score. To compute this score, the query trajectory is first mapped to the 2D polyline-based curvilinear coordinate system (as defined in Argoverse), wherein axes are defined to be tangential and perpendicular to a reference polyline. We define the alignment score to be the maximum tangential distance reached along the query trajectory (better alignment results in longer distances travelled along the reference polyline). To give a concrete example, if the maximum tangential distance for each of the candidate polylines is \textit{\{L1: 10 , L2: 25, L3: 2, L4: 20\}}, the sorted list of candidate polylines will be returned in the order $L_{a} = \textit{[L2, L4, L1, L3]}$.

    \item \textbf{Selecting Polylines}: We sort candidate polylines using two different methods of scoring because examining PIP score in isolation can sometimes be misleading. For example, a car moving slowly across an intersection could result in high PIP scores being assigned to polylines obtained from lanes with orthogonal directions of travel. Using the polyline-trajectory alignment score alone can also result in similar confusion. For example, a nearby protected turn lane that is parallel to the query trajectory's direction of travel may be assigned a high alignment score, even if the polyline represents a semantically different future.
    
    For this reason, it is important to rank polyline proposals using a combination of both metrics. In our implementation, we employ a heuristic-based selection process, wherein the polylines are drawn from the top of $L_{pip}$ and $L_{a}$ in alternating order. To give a concrete example, in the scenario we have posed above, querying for the best 2 polylines would return $L$ = \textit{[L4, L2]} (where \textit{L4} is the best PIP polyline and \textit{L2} is the best-aligned polyline).

    \item \textbf{Using Proposed Polylines}: During training, we query for and use only the top-ranked polyline proposal from each set. However, at inference time, it is possible to trade off prediction diversity and accuracy by controlling the the number of polyline proposals and predictions generated per polyline (e.g. 6 predictions conditioned on 1 polyline vs. 1 prediction conditioned on each of 6 polylines).
\end{enumerate}


\section{Improving Maps via Prediction}
\label{apdx:mapviaprediction}

\begin{wrapfigure}{r}{0.49\textwidth}
  \vspace{-10pt}
  \centering
  \begin{tabular}{ccc}
  \includegraphics[width=0.14\textwidth]{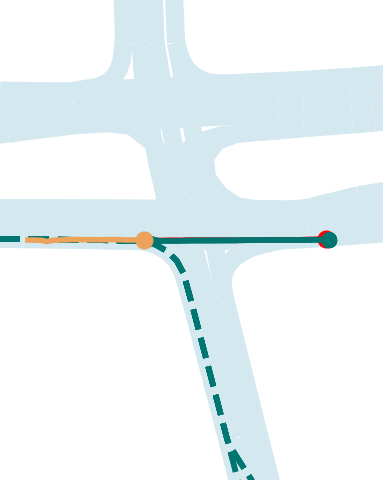} & \includegraphics[width=0.14\textwidth]{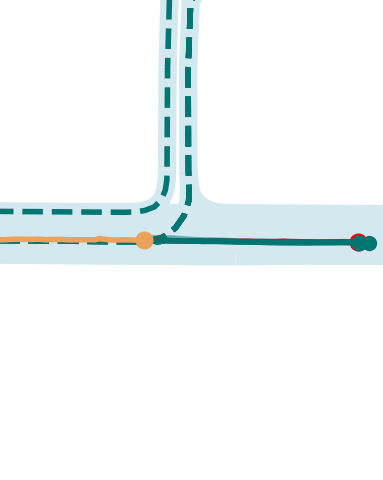} & \includegraphics[width=0.14\textwidth]{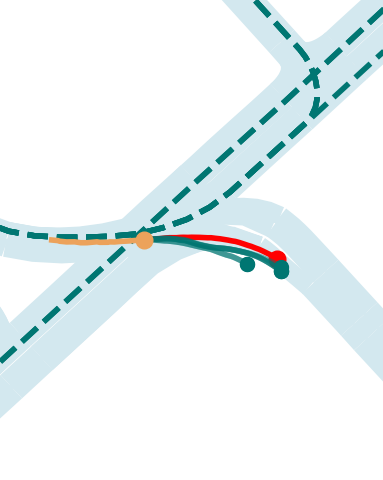}\\
  \end{tabular}
  \caption{BEV visualizations of three different intersections where accurate predictions based on polyline proposals from the vector map disagree with the observed mode of traffic behavior. Visualization style follows Fig. 3.}
  \vspace{-10pt}
  \label{fig:mapupdate}
\end{wrapfigure}

Planning and forecasting in AVs are tightly coupled to semantic map data that is collected, processed, and annotated offline. By examining for repeated and significant disagreement between heuristically chosen lane polylines and accurate forecasted trajectories, we can automatically identify map locations where the proposed lane polylines fail to capture the dominant modes of traffic behavior (shown in Fig. \ref{fig:mapupdate}).

This ancilary benefit of explicit path-conditioning could serve as an important feedback mechanism for generation and maintenance of safe and current maps. Without updates, maps can quickly become outdated in urban environments, as active construction and development modifies the road network and induces change in traffic patterns. One such way that map updates could be performed in an online setting is to assign a weighted prior for each map polyline, with value inversely proportional to the rate of disagreement between conditioned predictions and the corresponding polyline. These polyline weights can then be used as an input to a heuristic or learning-based polyline proposal module to enable dynamic selection of high-quality reference trajectories. As variables within the environment change (e.g. construction, weather, potholes), priors can be automatically updated to capture the updated distribution of driver behavior.

\section{Argoverse Challenge Entries}
\label{apdx:challengeresults}
Due to the dynamic nature of the Argoverse leaderboard, it can be difficult to accurately compare prediction performance against competing approaches during an ongoing competition. Rankings of state-of-the-art entries can shift rapidly and no information is available about individual entries; in the closing weeks of the competition, significant changes have been observed on a near-daily basis. Although WIMP out-performs all previously published and publicly-released methods in the evaluated MR and DE-based metrics, several of the most recent leaderboard entries have since improved further upon our results. Comparing against these entries, we report the state of the leaderboard as captured at two points in time: one on Mar. 23, 2020 (Table \ref{table:earlyleaderboard}) and the other on Jun. 8, 2020 (Table \ref{table:lateleaderboard}). 

\begin{table*}[h]
\begin{center}
\begin{small}
\begin{sc}
\begin{tabular}{lcccccc}
\toprule
Team Name & Rank & MR(K=6) & FDE(K=6) & ADE(K=6) & FDE(K=1) & ADE(K=1)\\
\midrule
``poly" & 1 & \textbf{0.14} & 1.50 & 0.91 & \underline{4.00} & \underline{1.79}\\
``lgn" \cite{liang2020learning} & 2 & \underline{0.16} & \textbf{1.36} & \textbf{0.87} & \textbf{3.78} & \textbf{1.71}\\
``lstm" & 3 & 0.17 & 1.67 & 0.99 & 4.26 & 1.91\\
``tests" & 5 & 0.18 & 1.50 & 0.93 & 4.44 & 2.02\\
``jean" & 6 & 0.18 & 1.48 & 0.93 & 4.17 & 1.86\\
``ust" & 7 & 0.19 & 1.45 & 0.92 & 4.09 & 1.86\\
``cxx" & 8 & 0.19 & 1.71 & 0.99 & 4.31 & 1.91\\
``uulm-mrm" & 9 & 0.22 & 1.55 & 0.94 & 4.19 & 1.90\\
``el camino" & 10 & 0.25 & 1.98 & 1.13 & 4.84 & 2.17\\
\midrule
WIMP ($M=6$) & 4 & 0.17 & \underline{1.42} & \underline{0.90} & 4.03 & 1.82\\
\bottomrule
\end{tabular}
\end{sc}
\end{small}
\end{center}
\caption{Motion forecasting performance evaluated on the Argoverse test set, reported for the top 10 (ranked by MR) entries on the Argoverse Forecasting Challenge leaderboard (as of May 23, 2020). MR and minimum FDE/ADE metrics are reported for both single ($K=1$) and multi-modal ($K=6$) prediction scenarios. Top-ranked entries have improved performance significantly beyond the bar set by 2019 Argoverse Forecasting Challenge joint-winners SAMMP~\cite{mercat2019multihead} and UULM. The best entry is bolded, while the second-best is underlined. WIMP ranks second on K=6 metrics for FDE/ADE and third for MR and K=1 metrics. }
\label{table:earlyleaderboard}
\end{table*}

\begin{table*}[]
\begin{center}
\begin{small}
\begin{sc}
\begin{tabular}{lcccccc}
\toprule
Team Name & Rank & MR(K=6) & FDE(K=6) & ADE(K=6) & FDE(K=1) & ADE(K=1)\\
\midrule
``jean" & 1 & \textbf{0.13} & 1.42 & 0.97 & \textbf{3.73} & \textbf{1.68}\\
``poly" & 2 & \underline{0.13} & 1.48 & 0.92 & 3.95 & 1.77\\
``alibaba-adlab" & 3 & 0.16 & 1.48 & 0.92 & 4.35 & 1.97\\
``lgn" \cite{liang2020learning} & 4 & 0.16 & \textbf{1.36} & \textbf{0.87} & \underline{3.78} & \underline{1.71}\\
``lstm" & 5 & 0.17 & 1.67 & 0.99 & 4.26 & 1.91\\
``ust" & 7 & 0.19 & 1.45 & 0.92 & 4.09 & 1.86\\
``cxx" & 8 & 0.19 & 1.71 & 0.99 & 4.31 & 1.91\\
``mt" & 9 & 0.22 & 1.66 & 0.98 & 8.23 & 3.58\\
``uulm-mrm" & 10 & 0.22 & 1.55 & 0.94 & 4.19 & 1.89\\
\midrule
WIMP ($M=6$) & 6 & 0.17 & \underline{1.42} & \underline{0.90} & 4.03 & 1.82  \\
\bottomrule
\end{tabular}
\end{sc}
\end{small}
\end{center}
\caption{Motion forecasting performance evaluated on the Argoverse test set, reported for the top 10 (ranked by MR) entries on the Argoverse Forecasting Challenge leaderboard (as of June 8 , 2020). MR and minimum FDE/ADE metrics are reported for both single ($K=1$) and multi-modal ($K=6$) prediction scenarios. Note that the leaderboard rankings have shifted significantly from Table \ref{table:earlyleaderboard}, with the addition of new entries and refinement of existing methods. WIMP ranks second for K=6 ADE/FDE metrics.}
\label{table:lateleaderboard}
\end{table*}

\section{Dynamic Visualizations}
\label{apdx:dynamicviz}
As trajectory prediction is a fundamentally three-dimensional task that requires integration of information across space and time, it can be difficult to capture temporal context using static 2D images alone. To address this issue, we provide dynamic visualizations (following the style of Fig. 3) for each of the BEV scenarios shown in the main text (Figs. 3-5). 

We also include additional dynamic visualizations from prediction scenarios that capture a broad range of interesting events: acceleration, braking, full stops, fast driving, exiting driveways, lane changes, left turns, right turns, wide turns, and use of protected turn lanes. These examples are intended to demonstrate that WIMP not only generates predictions that are accurate and diverse, but also generalizes to a wide variety of geographic and semantic settings. These visualizations will be made available here: \href{https://github.com/wqi/WIMP}{https://github.com/wqi/WIMP}.
\end{document}